\documentclass[runningheads]{llncs}

\usepackage{eccv}

\usepackage[dvipsnames]{xcolor}

\usepackage{epsfig}
\usepackage{graphicx}
\usepackage{amsmath}
\usepackage{amssymb}

\usepackage{multirow}
\usepackage{bm}
\usepackage{float}
\usepackage{colortbl}
\usepackage{color, xcolor} 
\usepackage{comment}
\usepackage{algorithm}
\usepackage{algorithmicx}
\usepackage{algpseudocode}
\usepackage{indentfirst}
\usepackage{array}
\usepackage{booktabs}
\usepackage{enumitem}

\usepackage{wrapfig}

\newcommand{\PreserveBackslash}[1]{\let\temp=\\#1\let\\=\temp}
\newcolumntype{C}[1]{>{\PreserveBackslash\centering}p{#1}}
\newcolumntype{R}[1]{>{\PreserveBackslash\raggedleft}p{#1}}
\newcolumntype{L}[1]{>{\PreserveBackslash\raggedright}p{#1}}

\definecolor{remark}{rgb}{1,.5,0} 
\definecolor{citecolor}{rgb}{0,0.443,0.737} 
\definecolor{linkcolor}{rgb}{0.956,0.298,0.235} 
\definecolor{gray}{gray}{0.5}
\definecolor{cyan}{rgb}{0.831,0.901,0.945}
\definecolor{teal}{rgb}{0,0.5,0.5}
\definecolor{lightskyblue}{rgb}{0.53,0.8,0.976}
\definecolor{llgray}{rgb}{0.85,0.85,0.85}

\usepackage{pifont}
\newcommand{\cmark}{\ding{51}}%
\newcommand{\xmark}{\ding{55}}%

\usepackage{lipsum}

\usepackage[normalem]{ulem}

\makeatletter\renewcommand\paragraph{\@startsection{paragraph}{4}{\z@}
  {.5em \@plus1ex \@minus.2ex}{-.5em}{\normalfont\normalsize\bfseries}}\makeatother

\newcolumntype{x}[1]{>{\centering\arraybackslash}p{#1pt}}
\newlength\savewidth

\newcommand{\lego}{LEGO\xspace}

\usepackage{eccvabbrv}

\usepackage{graphicx}
\usepackage{booktabs}

\usepackage[accsupp]{axessibility}  

\usepackage{hyperref}

\usepackage{orcidlink}

\usepackage[capitalize]{cleveref}
\crefname{section}{Sec.}{Secs.}
\Crefname{section}{Section}{Sections}
\Crefname{table}{Table}{Tables}
\crefname{table}{Tab.}{Tabs.}

\colorlet{dark-blue}{blue!70!black}
\colorlet{dark-green}{green!60!black}
\colorlet{dark-red}{red!80!black}
\definecolor{mypink}{RGB}{219, 48, 122}
\hypersetup{
 colorlinks=true,%
 citecolor=citecolor,%
 filecolor=citecolor,%
 linkcolor=linkcolor,%
 urlcolor=magenta
}

\makeatletter
\DeclareRobustCommand\onedot{\futurelet\@let@token\@onedot}
\def\@onedot{\ifx\@let@token.\else.\null\fi\xspace}

\def\eg{\emph{e.g}\onedot} 
\def\ie{\emph{i.e}\onedot} 
\def\cf{\emph{c.f}\onedot} 
\def\etc{\emph{etc}\onedot} 
 
\def\etal{\emph{et al}\onedot}
\makeatother

\begin{document}

\title{TreeSBA: Tree-Transformer for Self-Supervised Sequential Brick Assembly}

\titlerunning{Tree-Transformer for Self-Supervised Sequential Brick Assembly}
\author{Mengqi Guo* \and
Chen Li* \and
Yuyang Zhao \and
Gim Hee Lee
}

\authorrunning{Guo et al.}

\institute{Department of Computer Science, National University of Singapore
\\
\email{\{mengqi, gimhee.lee\}@comp.nus.edu.sg}
\\
\url{https://dreamguo.github.io/projects/TreeSBA}
}

\maketitle

\renewcommand{\thefootnote}{\fnsymbol{footnote}}
\footnotetext[1]{Equal contribution} 

\vspace{-3mm}
\begin{abstract}
    Inferring step-wise actions to assemble 3D objects with primitive bricks from images is a challenging task due to complex constraints and the vast number of possible combinations. Recent studies have demonstrated promising results on sequential LEGO brick assembly through the utilization of LEGO-Graph modeling to predict sequential actions. However, existing approaches are class-specific and require significant computational and 3D annotation resources.
    In this work, we first propose a computationally efficient breadth-first search (BFS) LEGO-Tree structure to model the sequential assembly actions by considering connections between consecutive layers. Based on the LEGO-Tree structure, we then design a class-agnostic tree-transformer framework to predict the sequential assembly actions from the input multi-view images. A major challenge of the sequential brick assembly task is that the step-wise action labels are costly and tedious to obtain in practice. We mitigate this problem by leveraging synthetic-to-real transfer learning. Specifically, our model is first pre-trained on synthetic data with full supervision from the available action labels. We then circumvent the requirement for action labels in the real data by proposing an action-to-silhouette projection that replaces action labels with input image silhouettes for self-supervision. Without any annotation on the real data, our model outperforms existing methods with 3D supervision by 7.8\% and 11.3\% in mIoU on the MNIST and ModelNet Construction datasets, respectively.

    \vspace{-2mm}
    \keywords{Sequential Assembly \and Self-Supervised \and Synthetic-to-Real}
    \vspace{-2mm}
\end{abstract}

\vspace{-4mm}
\section{Introduction}
\vspace{-2mm}
Assembling 3D objects from simple shape primitives is a common operation in the real world, such as house construction and furniture assembling. This task is even harder when only multi-view images instead of blueprints are available. 
Nonetheless, automatically generating the blueprint and assembling actions from images has great potential for real-world applications, especially for robotics, augmented and virtual reality, \etc.
Previous works~\cite{wu2020pq, zou20173d, paschalidou2019superquadrics} formulate the task as sequential part assembly by regarding each part of an object as a primitive. An intractably large set of primitives has to be defined in this case since objects consist of many different parts.
Recently, researchers have explored the task of \textbf{S}equential \textbf{B}rick \textbf{A}ssembly (\textbf{SBA})~\cite{Brick, LEGO, LEGO_graph}, which assembles different 3D shapes with the unit bricks as the primitives. 
Despite having finite primitives, the representation power of SBA is significantly large from the combinations of these primitives. For example, we can use only six $2\times 4$ \lego bricks to create 915,103,765 different combinations~\cite{eilers2016lego}.

\begin{figure}[t]
    \centering
    \vspace{-1mm}
    \includegraphics[width=.95\linewidth]{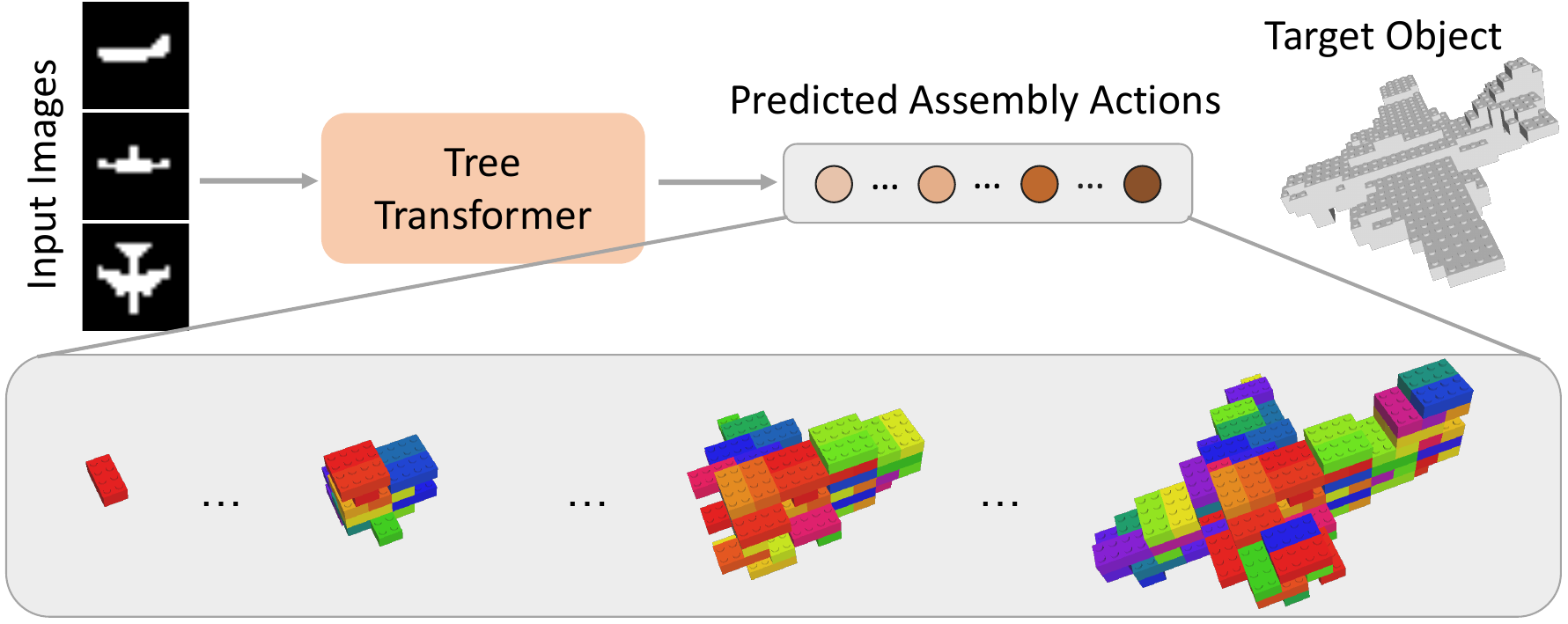}
    \vspace{-3mm}
    \caption{Example of self-supervised SBA. The 3D object is assembled from \lego bricks based on the predicted sequence of actions.
    Brick color is for better visualization.} 
    \vspace{-6mm}
    \label{fig:first}
\end{figure}

In this paper, we study the SBA task to predict the step-wise actions for the assembly of a 3D object from a set of multi-view images using \lego bricks as the shape primitive. The state-of-the-art work on SBA~\cite{Brick} leverages graph modeling to predict the sequential action and optimize it via reinforcement learning. 
However, each brick is compared to all previous bricks to determine whether there is a connection. Consequently, the time complexity increases tremendously with the number of bricks. Moreover, the model in \cite{Brick} is class-specific, \ie one model for each class, and requires 3D voxel supervision.

In view of the above-mentioned limitations, we design a class-agnostic tree transformer for the task of SBA. Specifically, we first design a BFS LEGO-Tree to represent the sequential assembly actions. 
In contrast to the GNNs that connect new bricks with all previous bricks in \cite{Brick}, our LEGO-Tree only considers the connection type between two consecutive layers and thus significantly reducing the computational costs. 
Based on the tree structure, we further propose a class-agnostic tree-transformer with both tree and global positional embeddings to learn the construction. Given that the ground truth action labels for the assembly sequences are expensive and tedious to obtain in practice, we leverage \textit{synthetic-to-real transfer learning} to circumvent the requirement for annotating large amounts of ground truth action labels. 
Specifically, we first pre-train our model with synthetic data where ground truth action labels can be synthetically generated in large quantities for full supervision. The pre-trained model is then further fine-tuned on the real data, where we alleviate the lack of ground truth action labels by proposing an action-to-silhouette projection that replaces action labels with image silhouettes for self-supervision. 
Moreover, we also introduce a tree augmentation technique to further improve the generalization capacity of our model to real data.

We validate our proposed approach on both synthetic and real data, where we achieve better results with less computational costs. Without any annotations on the real data, our model outperforms state-of-the-art methods with 3D supervision by mIoU of 7.8\% and 11.3\% on the MNIST and ModelNet Construction datasets, respectively. \textbf{Our contributions} are as follows: 
1) We design the computational efficient BFS LEGO-Tree to represent the sequential assembly actions.
2) We propose the class-agnostic tree-transformer to encode both tree and global information for SBA. 3) We introduce a self-training scheme, \ie action-to-silhouette projection to facilitate transfer learning from synthetic-to-real data and achieve the state-of-the-art performance on the task of self-supervised SBA.

\section{Related Work}

\vspace{-1mm}
\paragraph{3D Reconstruction.} 
3D reconstruction of objects from a set of multi-view images is an extensively studied problem in computer vision because of the wide applications in robot navigation, scene understanding, 3D modeling, animation, \etc. Traditional methods \cite{debevec1996modeling, curless1996volumetric, snavely2006photo} are based on geometry-based photogrammetry and have achieved reasonable quality 3D reconstructions. With the rapid advancement in deep learning and the emergence of large 3D object datasets, \eg ShapeNet~\cite{chang2015shapenet} and ModelNet~\cite{wu20153d}, increasingly more deep learning-based approaches~\cite{tulsiani2018multi, xie2019pix2vox, fan2017point, GernotRiegler2016OctNetLD, tatarchenko2017octree, choy20163d, kar2017learning, LiJiang2018GALGA, wu2016learning, arsalan2017synthesizing, mandikal20183d, XingyuanSun2018Pix3DDA, wu2017marrnet, ZaiShi20213DRETRES, park2019deepsdf, mildenhall2020nerf} are proposed to recover 3D geometry from images. However, the task of 3D reconstruction aims to recover geometry information from images while we aim to infer the spatial semantic information on the connections and the assembly sequence of the construction.

\vspace{-1mm}
\paragraph{Part Assembly.}
Part assembly is a well-study task, and existing works ~\cite{paschalidou2019superquadrics, tulsiani2017learning, gadelha2020learning} have successfully tackled the primitive assembly task using deformable primitives. Some works extract 3D assembled information, \eg 6d pose of parts using convolutional neural networks ~\cite{li2020learning, tulsiani2018factoring}, recurrent neural networks ~\cite{zou20173d, wu2020pq}, and graph modeling ~\cite{zhan2020generative}. Other works formulate the assembly task as a shape-fitting problem ~\cite{zakka2020form2fit}, shape-matting problem ~\cite{chen2022neural, wu2023leveraging} or part retrieval and assembly problem ~\cite{xu2023unsupervised}.
Although those methods provide precise results in optimizing the part shape and predicting the pose and location of parts, the connection information is missing. 
On the other hand, the sequential brick assembly task provides another way to assemble objects with unit primitives given pre-defined connection types and geometry constraints.

\vspace{-1mm}
\paragraph{Sequential Brick Assembly.}
Most previous works on SBA ~\cite{LEGO, gower1998lego, lennon2021image2lego, lee2015finding, LEGO_graph, Brick} require 3D voxel labels and cannot directly predict action sequences from images.
Kim \etal~\cite{LEGO} generate action sequence from the input 3D voxel based on Bayesian optimization~\cite{brochu2010tutorial}. Thompson \etal~\cite{LEGO_graph} utilize Deep Generative Models of Graph (DGMG)~\cite{li2018learning} to assemble a \lego object from category information. 
Recently, Wang \etal~\cite{wang2022translating} predict action sequence by translating a set of detailed manuals from images. However, the manual image for each step is required as supervision.
Walsman \etal~\cite{walsman2022break} propose a transformer-based model to predict action sequences from images along with an interactive simulator. Nevertheless, they rely on detailed 3D models with conjunction information 
and step-wise images generated from an interactive environment.
The closest setting to our work is B$^3$ \etal~\cite{Brick}, which improves upon DGMG~\cite{li2018learning} with a reinforcement learning (RL) framework~\cite{schulman2017proximal}.
However, it takes a long time for B$^3$ to converge even when the number of bricks is small due to the large search space.
Furthermore, B$^3$ is a class-specific model and requires 3D labels during training.
In this paper, we design a class-agnostic tree transformer for the task of SBA. Without any labels for the realistic dataset, our two-stage pipeline achieves better performance with less training time compared with existing GNN-based methods.

\section{Our Method}

\begin{figure*}[t]
\centering
\vspace{-1mm}
\includegraphics[width=1.0\linewidth]{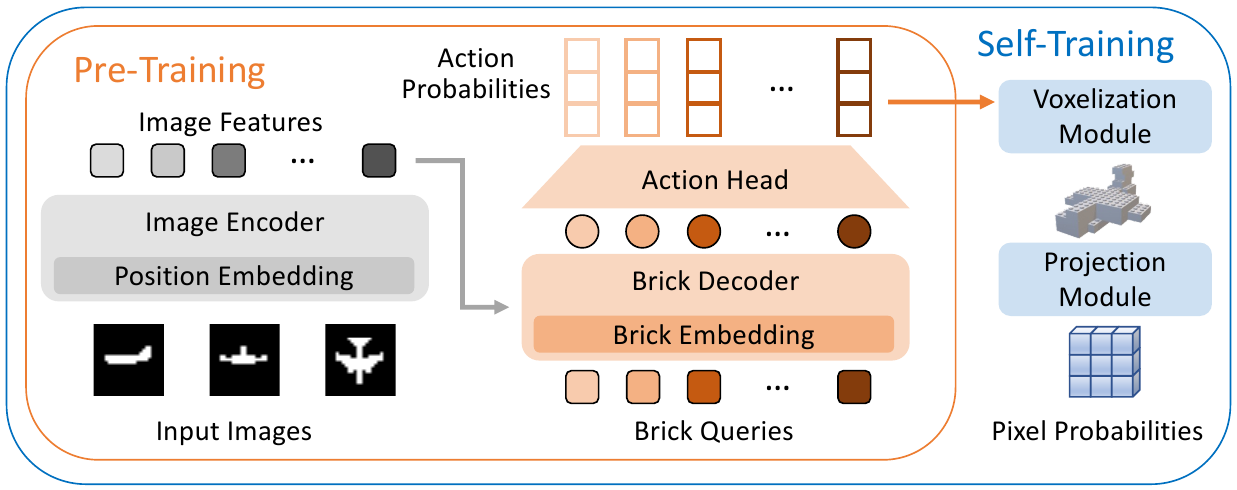}
\vspace{-6mm}
\caption{Overview of our two-stage framework that consists of a pre-training stage and a self-training stage. 
In the pre-training stage, we use an image encoder to extract the image features, which are fed into a brick decoder to estimate the action sequence. The estimated actions are supervised with ground truth action labels. In the self-training stage, we use voxelization and projection modules to transfer action probabilities into pixel probabilities such that the model can be self-supervised with the input images.
}
\vspace{-4mm}
\label{fig:framework}
\end{figure*}

\vspace{-1mm}
\paragraph{Problem Definition.}
Given a set of $S\times S$ silhouette images $I=[I_0,I_1,I_2]$ captured from a target 3D object, our model aims to predict a brick-wise action sequence $A=[a_0,a_1,\cdots,a_N]$ that assembles a set of \lego bricks $B=[b_0,b_1,\cdots,b_N]$ into the 3D object, where $N$ denotes the maximum number of actions and bricks. Each action $a_n \in \{0,1,\cdots,C\}$ represents the connections between the brick $b_n$ and all its neighbors. 

\vspace{-1mm}
\paragraph{Action Definition.}
The connection type between two \lego bricks is divided into four categories: x-axis offset, y-axis offset, z-axis offset and relative rotation. We define the z-axis offset separately as direction type $dir_n$ and the remaining as tree connection type $t_n$. 
Specifically, $dir_n \in \{0,1\}$, where $0$ represents that the brick $b_n$ is on top of its parent brick, and $1$ indicates the opposite direction, \ie at the bottom of its parent brick. 
$t_n \in \{0,1,...,T\}$ is the different combination of the x-axis offset, y-axis offset, and relative rotation. 
We define the action as $a_n=[a_n^{up} (dir_n=0), a_n^{down} (dir_n=1)]$, where 
$a_n^{up}$ ($a_n^{down}$) is the sub-action representing the top (bottom) of the current brick. Each sub-action represents the connection type between the current brick to all its neighboring bricks. 

\vspace{-1mm}
\paragraph{Overview.}
Our overall framework is illustrated in Fig.~\ref{fig:framework}.
We design a transformer-based model for the task of SBA, where the inputs are the multi-view images of the 3D object, and the outputs are the step-wise actions to construct the 3D object. The assembly actions are represented by the BFS \lego-Tree to encode connection information. To address the limited generalization ability and avoid the annotation cost, we leverage \textit{synthetic-to-real transfer learning} to propose a two-stage framework. Specifically, we first pre-train the model on the ground truth action labels of the synthetic data generated by random assembly of the \lego bricks.
The pre-trained model is then fine-tuned on the real data by self-training. To facilitate the self-training, we first convert the output action probabilities to 3D voxels and then project them to the input camera views. With this action-to-silhouette projection, the whole training process can be self-supervised with the input images.

\vspace{-1mm}
\subsection{\lego-Tree modeling}
\label{LEGO_Tree}
\begin{figure}[t]
\centering
\includegraphics[width=.70\linewidth]{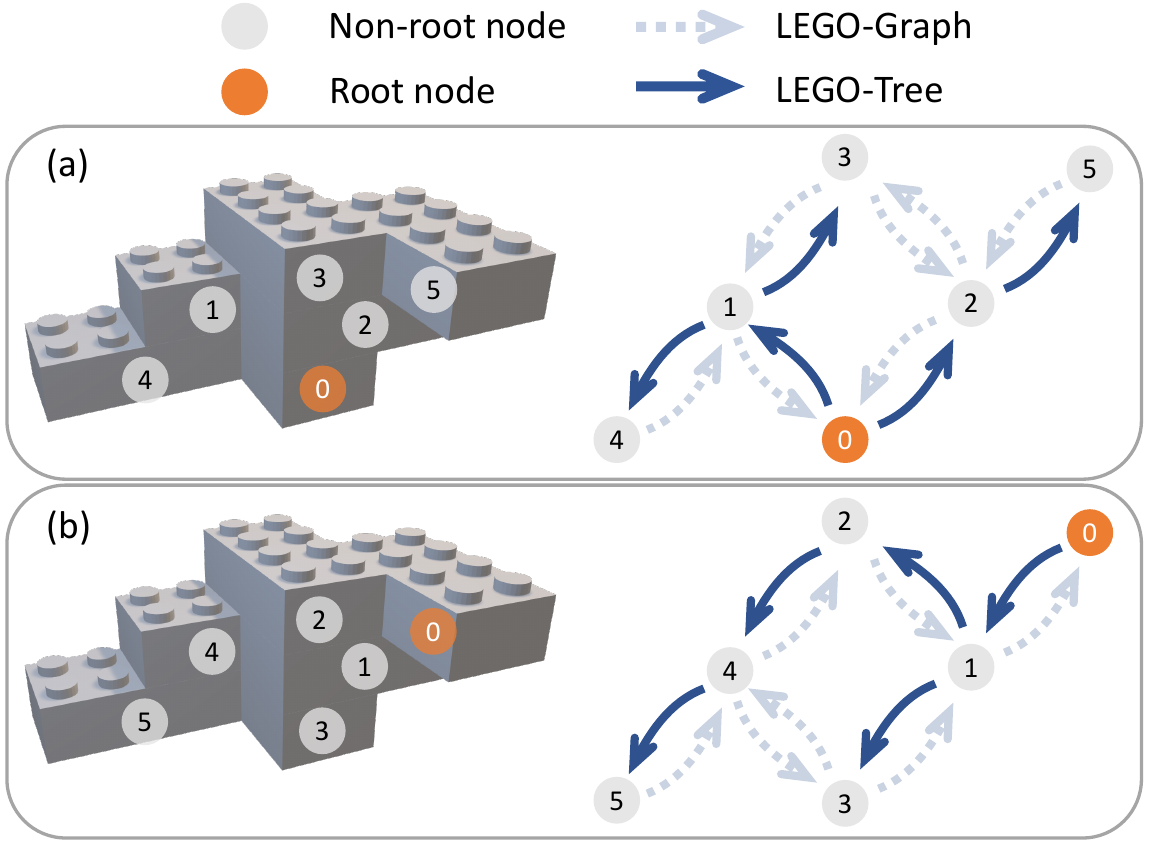}
\vspace{-2mm}
\caption{Illustration of the \lego-Tree model and action reordering data augmentation. Node number represents the index of action in the sequence and the index of brick in the tree. (a) shows a BFS \lego-Tree (solid line) generated from a \lego-Graph (dotted line). (b) is an action-reordering augmented sample of (a).}
\vspace{-4mm}
\label{fig:LEGOTree}
\end{figure}

Previous works~\cite{Brick, LEGO_graph} commonly view a \lego object as a graph (\lego-Graph), where the nodes and edges represent bricks and connection type, respectively. 
When a new brick is added, the connection type between this brick and all previous $n$ bricks needs to be re-predicted. This is computationally costly with a complexity of $O(n^2)$. We thus take a different perspective and predict the connection type of one brick and all its neighbors in one 
go. We design a BFS \lego-Tree to represent the construction process as shown in Fig.~\ref{fig:LEGOTree}(a). 
Given the root brick $b_0$, we first predict an action $a_0$, which represents connecting to both two neighbor bricks $b_1$ and $b_2$. We can then construct the whole object by predicting the action type for all subsequent bricks in the order of the BFS algorithm.
Consequently, the computation complexity is reduced to $O(n)$. Note that we normalize the root brick at the center of the 3D voxel space to eliminate the spatial bias.

\vspace{-1mm}
\subsection{Tree-Transformer}
\label{Tree-Transformer}
We propose a transformer-based encoder-decoder model, \ie, tree-transformer, to handle the sequential assembly task. There are three main components in the tree-transformer: image encoder to extract image features, brick decoder to predict the brick feature sequentially, and action head to predict the action probability. Specifically, we leverage the ViT backbone \cite{dosovitskiy2020vit} as the encoder and extract features, which are used as the keys and values in the brick decoder. To aggregate information and predict action for each brick, we design the brick queries based on tree structure and global geometry information. The action prediction of the tree-transformer is formulated as:
\begin{equation}
\label{eq:decoder}
\begin{aligned}
P^{A}=f_{head} \Bigl( f_{dec} \bigl( f_{enc}(I),E(B) \bigr) \Bigr).
\end{aligned}
\end{equation}
$P^A=[p_0^A,p_1^A,\cdots,p_N^A]$, where $p_n^A$ is a $C$-dimensional vector for the predicted actions probability, $f_{enc}(\cdot)$, $f_{dec}(\cdot)$, and $f_{head}(\cdot)$ are the transformer-based encoder, decoder, and MLP modules, respectively. $E(B)$ is the query for all the bricks, including tree structure embedding and global geometry embedding.

\paragraph{Tree Structure Embedding.}
One brick can be defined by the connection and location information in the tree, which can be represented by three components: the connection type, connection direction, and depth location of the brick in the tree. The tree structure embedding is thus given by:
\begin{equation}
\begin{aligned}
E_{str}(b_n)=E_{t}(t_n)+E_{dir}(dir_n)+E_{dep}(dep_n),
\end{aligned}
\end{equation}
where $E_{t}(\cdot)$, $E_{dir}(\cdot)$ and $E_{dep}(\cdot)$ are the embedding layers of the connection type, connection direction, and depth location, respectively. $t_n$ is the tree connection type that encodes information on the $x, y, z$-axis offsets from parent brick to brick $n$. $dir_n$ denotes the direction type from parent brick to brick $n$, and $dep_n \in \{0,1,\cdots,N\}$ is the depth of brick $n$ in the \lego-Tree model. The $t_n, dir_n, dep_n$ are all computed from the previous predicted assembly actions $a_{0:(n-1)}$.

\paragraph{Global Geometry Embedding.}
Tree structure alone cannot fully represent a brick. As shown in Fig.~\ref{fig:LEGOTree}(a), two different bricks 3 and 5 share the same tree structure embedding, which makes them indistinguishable. 
In view of this problem, we further introduce the global location information for the $n$th brick:
\begin{equation}
\begin{aligned}
E_{geo}(b_n)=E_{p}(x_n)+E_{p}(y_n)+E_{p}(z_n)+E_{r}(r_n),
\end{aligned}
\end{equation}
where $x_n, y_n, z_n \in \{0,1,\cdots,S\}$ are the global coordinates of the brick, $r_n \in \{0,1\}$ is the global orientation, and $E_{p}(\cdot)$ and $E_{r}(\cdot)$ are position and orientation embedding layers, respectively. With the global position and orientation information, the model can easily discriminate bricks with the same patterns in the \lego-Tree. 
$x_n, y_n, z_n, r_n$ are all computed from the previous predicted assembly actions $a_{0:(n-1)}$.

\paragraph{Brick Query.}
Finally, we combine the tree structure embedding and global location embedding to form the query for each brick $b_n$:
\begin{equation}
\begin{aligned}
E(b_n)=E_{str}(b_n)+E_{geo}(b_n).
\end{aligned}
\end{equation}
Given the queries for all 
bricks $E(B)=[E(b_0),E(b_1),\cdots,E(b_N)]$, we can decode and predict the corresponding actions for the \lego assembly with Eqn.~\eqref{eq:decoder}.

\vspace{-1mm}
\paragraph{Brick Collision Avoidance.}
To ensure the buildability of predicted assembly actions, we employ a binary occupancy voxel during the assembly process to prevent brick collisions. Any predicted action that results in an overlap with the occupancy voxel is deemed illegal and subsequently disregarded.
This collision avoidance mechanism is consistently applied in both training and testing phases, guaranteeing 100\% legality for predicted actions. Notably, B$^3$~\cite{Brick} maintains 90\% legality rate. Despite this strict approach to collision avoidance, our model outperforms baselines due to the robust learning ability of the Tree-Transformer.

\begin{figure}[t]
\centering
\vspace{-2mm}
\includegraphics[width=.9\linewidth]{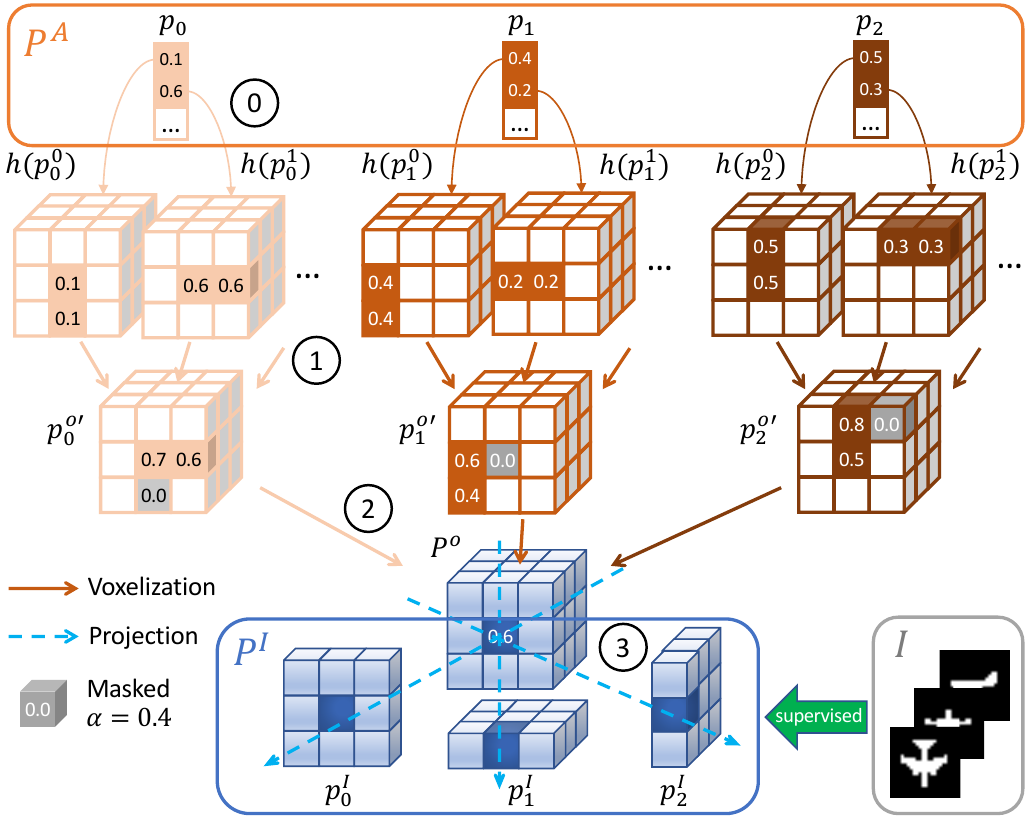}
\vspace{-3mm}
\caption{
An example to illustrate the silhouette self-training with the input images $I$. Given the action probabilities $P^A=[p_0,p_1,p_2]$, a 4-step approach is used to compute the pixel probabilities $P^I=[p^I_0,p^I_1,p^I_2]$. Refer to Sec.~\ref{sec:self-training} for more details.
} 
\vspace{-6mm}
\label{fig:projection}
\end{figure}

\vspace{-1mm}
\subsection{Pre-training on Synthetic Data}
\label{Pretrain}
\paragraph{Action Loss.}
We first train the tree-transformer with synthetic data that contains ground truth labels for each action. Since we view each action component as a classification task, the action loss is defined as cross entropy:
\begin{equation}
\begin{aligned}
\mathcal{L}_{act} = \sum_{i=n}^N - y_{n} \log p_n^A,
\end{aligned}
\end{equation}
where $y_n$ is the action label of brick $n$. We further propose a data augmentation technique to improve the generalization ability for unseen realistic objects.

\vspace{-1mm}
\paragraph{Action Reordering.}
The root brick has to be selected prior to building the BFS \lego-tree. By default, the root brick is at the bottom of the \lego object as illustrated in Fig.~\ref{fig:LEGOTree}(a). However, this is not always true for most generic objects, \eg some thin structures cannot be assembled with the root brick at the bottom. 
We thus augment the synthetic data by randomly selecting one brick as the root brick and putting it at the center of the 3D space. Consequently, the structure of the \lego-Tree and the sequential actions are also changed accordingly. We show an example of the action reordering in Fig.~\ref{fig:LEGOTree}(b). For more details 
on the action reordering, please refer to the supplementary material.

\vspace{-1mm}
\subsection{Self-training on Real Data} 
\vspace{-1mm}
\label{sec:self-training}
We further fine-tune the pre-trained model on real data to bridge the synthetic-to-real gap and thus improve the generalization capacity. Given that action-level and voxel-level labels are not always available, we propose to leverage the input multi-view images for self-training.
Specifically, we propose two probabilistic transformation modules to convert the output actions into silhouettes.
As shown in Fig.~\ref{fig:projection}, we first convert the output action probabilities to 3D voxel space and then project them to the input camera views. The projected images are used in the silhouette loss to optimize the model.

\vspace{-1mm}
\paragraph{Probabilistic Voxelization.}
We first convert the output action probabilities into the occupancy probabilities in the 3D voxel space. Specifically, each action type has a position offset relative to the parent brick and hence we can construct voxel probabilities by computing the global position of each brick along the \lego-Tree. This process is represented by a mapping function $h(\cdot)$ in \textbf{Step 0} of Fig.~\ref{fig:projection}.
There are two ways to handle the case where multiple actions correspond to one voxel: summing over all the actions and selecting the most confident actions. Intuitively, the most confident action is the predicted action for this step, which is a good choice to represent occupancy. However, it ignores the penalty or reward of the other actions which makes the model hard to converge. 
Consequently, we sum over all the action probabilities as the occupancy probabilities of the voxels occupied by the corresponding brick in \textbf{Step 1} of Fig.~\ref{fig:projection}:
\begin{equation}
p^{o\prime}_n = \sigma \left( \sum_{c} h(p_n^A), \alpha \right),
\label{eq:map+sum+mask}
\end{equation}
where $\sigma(\cdot)$ is a masking function that filters out the value lower than the masking threshold $\alpha$.
Note that $p^{o\prime}_n$ is the occupancy probabilities obtained from one brick, while each voxel can be occupied by more than one action. Thus, we average over all the masked actions to obtain the global occupancy probabilities $P^o$ in \textbf{Step 2} of Fig.~\ref{fig:projection}:
\begin{equation}
P^o=\delta_n(p^{o\prime}_0,p^{o\prime}_1,\cdots,p^{o\prime}_n).
\label{eq:average_voxelization}
\end{equation}
$\delta_n(\cdot)$ 
gives the average of non-zero values over $n$ actions.

\vspace{-1mm}
\paragraph{Probabilistic Projection.}
Given the voxelized probabilities $P^o$, we project them along the ray direction of the input camera view in \textbf{Step 3} of Fig.~\ref{fig:projection}:
\begin{equation}
p^I_v=\delta_v(P^o),
\label{eq:average_projection}
\end{equation}
where $\delta_v(\cdot)$ is a parallel projection function along the $v$ ray direction and $p^I_{v}$ is the projected silhouette image. The silhouette predictions are denoted as $P^I = [p^I_0, p^I_1, p^I_2]$.

\vspace{-1mm}
\paragraph{Silhouette Loss.} 
Finally, the model is supervised by the L2 distance between the input images and projected images:
\begin{equation}
\mathcal{L}_{sil}=||I-P^I||^2.
\label{eq:rec_loss}
\end{equation}
$I$ and $P^I$ are the input and projected images, respectively.

\vspace{-1mm}
\section{Experiments}
\vspace{-1mm}
\subsection{Experimental Setup}
\vspace{-1mm}

\begin{table*}[t]
\centering{
\small
\setlength{\tabcolsep}{0.03cm}
\caption{\textcolor{black}{Comparison with the state-of-the-art methods on the MNIST-C dataset. ``Pre-train'' represents the dataset used during the synthetic pre-training stage and the ``MNIST-C''  denotes training with or without MNIST-C and the type of supervision. Rows shaded in \textcolor{gray}{Lightgray} denote the class-specific models while unshaded rows denote the class-agnostic models.}}
\label{Tab:MNIST}
\vspace{-3mm}
\begin{tabular}{l|cc|cccccccccc|c}
\toprule
\multirow{2}{*}{Method} & \multicolumn{2}{c|}{Train Dataset} &\multicolumn{10}{c|}{IoU} & \multirow{2}{*}{mIoU} \\
\cmidrule{2-13}
& Pre-train & MNIST-C & \textit{0} & \textit{1} & \textit{2} & \textit{3} & \textit{4} & \textit{5} & \textit{6} & \textit{7} & \textit{8} & \textit{9} & \\
\midrule
\rowcolor{llgray}
BO~\cite{LEGO} & \textcolor{dark-red}{\xmark} & Voxel & $51.0$ & $70.0$ & $48.0$ & $38.0$ & $51.0$ & $38.0$ & $51.0$ & $55.0$ & $51.0$ & $52.0$ & $50.5$ \\
\rowcolor{llgray}
B$^3$~\cite{Brick} & \textcolor{dark-red}{\xmark}  & Voxel & $52.0$ & $74.0$ & $48.0$ & $50.0$ & $56.0$ & $45.0$ & $60.0$ & $66.0$ & $50.0$ & $60.0$ & $56.1$ \\
\rowcolor{llgray}
Ours & RAD-S & Image & $\underline{59.7}$ & $\underline{78.5}$ & $58.3$ & $58.4$ & $\textbf{67.9}$ & $\textbf{56.8}$ & $\textbf{63.3}$ & $71.1$ & $\underline{61.5}$ & $\underline{66.0}$ & $\textbf{64.2}$ \\
\midrule
Ours & RAD & \textcolor{dark-red}{\xmark} & $39.9$ & $74.9$ & $46.2$ & $49.7$ & $46.8$ & $36.2$ & $41.5$ & $57.0$ & $46.8$ & $55.8$ & $49.5$ \\
Ours & RAD-S & \textcolor{dark-red}{\xmark}  & $53.0$ & $74.6$ & $51.6$ & $50.2$ & $59.5$ & $46.4$ & $60.3$ & $59.5$ & $55.3$ & $58.5$ & $56.9$ \\
Ours & RAD & Image & $\textbf{62.0}$ & $75.9$ & $\underline{59.8}$ & $\underline{60.2}$ & ${60.4}$ & ${54.8}$ & $\underline{61.8}$ & $\textbf{72.3}$ & ${60.1}$ & ${65.8}$ & ${63.3}$ \\
Ours & RAD-S & Image & $58.6$ & $\textbf{77.8}$ & $\textbf{59.9}$ & $\textbf{60.7}$ & $\underline{63.0}$ & $\textbf{56.8}$ & ${61.4}$ & $\underline{72.0}$ & $\textbf{61.6}$ & $\textbf{67.4}$ & $\underline{63.9}$ \\
\bottomrule
\end{tabular}}
\vspace{-5mm}
\end{table*}

Following~\cite{Brick}, we conduct experiments on the Randomly Assembled Object Construction Dataset (RAD), MNIST Construction Dataset (MNIST-C), and ModelNet Construction Dataset (ModelNet-C).

\vspace{-1mm}
\paragraph{RAD \& RAD-S \& RAD-1k.}
RAD-1k~\cite{Brick} is a synthetic 3D \lego dataset built with $2\times4$ bricks. Each brick has eight studs and fit cavities, and thus there are 16 connection types in total. RAD-1k~\cite{Brick} contains 800 objects for training and 200 objects for testing, and each object is built with 15 bricks randomly with a predefined connection configuration. 
To fulfill the assembly requirements in the real world, we follow the connection configuration but relax the constraint on the connection direction to allow the new brick to be connected at the bottom of the previous brick. As a result, the total number of connection types between two $2\times4$ bricks is expanded from 16 to 32. In addition, we increase the brick number for each object to 20 for RAD-S, and 60 for RAD to better align with realistic scenarios.

\vspace{-1mm}
\paragraph{MNIST-C.}
MNIST-C~\cite{Brick} is a 3D \lego dataset by expanding along the channel dimension of 2D images from the MNIST dataset. For each class, there are 400 samples for training and 100 samples for testing. There are only six connection types for this dataset, but we still predict all 32 connection types to be consistent with the synthetic dataset.
The number of occupied voxels for 3D objects is around 140, which requires about 20 bricks for construction.

\vspace{-1mm}
\paragraph{ModelNet3-C \& ModelNet40-C.}
ModelNet-C~\cite{Brick} is a 3D \lego dataset by converting the voxels into \lego bricks from the ModelNet~\cite{wu20153d} dataset. 
Only 3 classes (airplane, monitor, and table) are used in \cite{Brick}, which we denoted as ModelNet3-C in this paper, and there are 1,477 samples for training and 299 samples for testing. We extend the construction dataset to all 40 classes of the ModelNet40~\cite{wu20153d} into ModelNet40-C, which contains 9,787 training samples and 2,452 testing samples.
To reduce the impact of scale, we re-scale the 3D objects so that the number of occupied voxels is around 320, which requires about 60 bricks for construction. 

\vspace{-1mm}
\paragraph{Implementation Details.}
We use the 4-layer ViT~\cite{dosovitskiy2020vit} for the image encoder and the 4-layer vanilla transformer deocder~\cite{vaswani2017attention} with a brick embedding module as the brick decoder.
The model is trained end-to-end with a batch size of 32 and Adam optimizer with weight decay $5e^{-4}$ is adopted. Learning rate is set to $1e^{-3}$ for synthetic pre-training, $1e^{-4}$ for the MNIST-C self-training, and $1e^{-5}$ for the ModelNet-C self-training. 
The model is pre-trained on RAD for 1000 epochs and fine-tuned on the MNIST-C and ModelNet-C for 40 and 10 epochs, respectively.
The voxelization masking threshold is $\alpha=0.4$. 
The model is trained and evaluated on a single Tesla V100 GPU.

\vspace{-1mm}
\paragraph{Evaluation Metric.} 
The performance is evaluated based on the similarity between the assembled and the ground truth 3D objects. 
We build the \lego object with the predicted action sequence with \lego bricks and convert it into voxel space. 
Intersection-over-union (IoU) between the occupied voxels of the assembled and the ground truth objects is adopted as the evaluation metric:
\begin{equation}
    \text{IoU}(C,T) = \frac{C \cap T}{C \cup T},
\end{equation}
where $C$ and $T$ denote the voxel representation of the assembled object and the ground truth object, respectively.

\vspace{-1mm}
\subsection{Comparison with state-of-art methods}
\vspace{-1mm}
\paragraph{Baselines.}
We first compare our framework with three benchmarks on SBA: Bayesian Optimization-based approach (BO)~\cite{LEGO}, Supervised Learning method (SL)~\cite{LEGO_graph}, and Brick-by-Brick (B$^3$)~\cite{Brick}. 
We also compare to the typical part assembly method SuperQuadric (SQ)~\cite{paschalidou2019superquadrics}. 
SL requires action level annotation thus we only show results on the RAD-1k dataset, while SQ requires 3D point cloud annotation thus we only show results on the ModelNet3-C dataset. 
In comparison, our model is pre-trained on the synthetic RAD dataset and fine-tuned on realistic data via self-training, which does not require any annotation for realistic data. 

\begin{table}[t]
\centering
\small
\setlength{\tabcolsep}{0.18cm}
\caption{\textcolor{black}{Comparison with the state-of-the-art methods on ModelNet3-C. ``Pre-train'' represents the dataset our model used in the synthetic pre-training stage and ``ModelNet3-C'' denotes training with or without ModelNet3-C and the type of supervision. Rows shaded in \textcolor{gray}{Lightgray} denote the class-specific models while unshaded rows denote the class-agnostic models.}}
\label{Tab:ModelNet}
\vspace{-3mm}
\resizebox{0.7\linewidth}{!}{\begin{tabular}{l|cc|ccc|c}
\toprule
\multirow{2}{*}{Method} & \multicolumn{2}{c|}{Train Dataset} &\multicolumn{3}{c|}{IoU} & \multirow{2}{*}{mIoU} \\
\cmidrule{2-3}\cmidrule{4-6}
& Pre-train & ModelNet3-C & \textit{airplane} & \textit{monitor} & \textit{\ \ \ table\ \ \ } & \\
\midrule
\rowcolor{llgray}
SQ~\cite{paschalidou2019superquadrics} & \textcolor{dark-red}{\xmark} & PointCloud & $27.1$ & $40.2$ & $20.3$ & $29.2$ \\
\rowcolor{llgray}
BO~\cite{LEGO} & \textcolor{dark-red}{\xmark} & Voxel & $25.0$ & $47.0$ & $17.0$ & $29.7$ \\
\rowcolor{llgray}
B$^3$~\cite{Brick} & \textcolor{dark-red}{\xmark} & Voxel & $36.0$ & $30.0$ & $\textbf{25.0}$ & $30.3$ \\
\rowcolor{llgray}
Ours & RAD & Image & $\textbf{44.6}$ & $\textbf{59.9}$ & $\underline{23.5}$ & $\textbf{42.7}$ \\
\midrule
Ours & RAD & \textcolor{dark-red}{\xmark} & $34.3$ & $55.1$ & $16.1$ & $35.1$ \\
Ours & RAD & Image & $\underline{42.2}$ & $\underline{59.5}$ & $23.4$ & $\underline{41.6}$ \\
\bottomrule
\end{tabular}}
\vspace{-2mm}
\end{table}

\begin{table*}[t]
\centering{
\small
\setlength{\tabcolsep}{0.02cm}
\caption{\textcolor{black}{Results on ModelNet40-C. Due to limited space, we only show the IoU of partial classes and the mIoU over all 40 classes.}}
\vspace{-3mm}
\label{Tab:MoreModelNet}
\begin{tabular}{cc|ccc|ccccccc|c}
\toprule
\multicolumn{2}{c|}{Train Dataset} & \multicolumn{10}{c|}{IoU} & \multirow{2}{*}{mIoU} \\
\cmidrule{1-2}\cmidrule{3-12}
SP & Self-training & \textit{airplane} & \textit{monitor} & \textit{table} & \textit{person} & \textit{piano} & \textit{sink} & \textit{sofa} & \textit{stairs} & \textit{stool} & \textit{toilet} & \\
\midrule
RAD & \textcolor{dark-red}{\xmark} & $34.3$ & $55.1$ & $16.1$ & $42.9$ & $40.9$ & $31.2$ & $38.1$ & $31.2$ & $23.0$ & $49.3$ & $36.5$ \\
RAD & Mod3-C & $\textbf{42.2}$ & $\textbf{59.5}$ & $\underline{23.4}$ & $\underline{45.9}$ & $\textbf{49.1}$ & $\underline{38.9}$ & $\underline{53.9}$ & $\textbf{40.1}$ & $\underline{28.7}$ & $\underline{54.3}$ & $\underline{45.1}$ \\
RAD & Mod40-C & $\underline{38.1}$ & $\underline{59.1}$ & $\textbf{23.7}$ & $\textbf{46.8}$ & $\underline{48.7}$ & $\textbf{40.4}$ & $\textbf{54.8}$ & $\underline{38.8}$ & $\textbf{30.5}$ & $\textbf{55.2}$ & $\textbf{45.4}$ \\
\bottomrule
\end{tabular}}
\vspace{-2mm}
\end{table*}

\vspace{-1mm}
\paragraph{MNIST-C.} 
We compare our models under different settings with BO~\cite{LEGO} and B$^3$~\cite{Brick} in Tab.~\ref{Tab:MNIST}. We make the following observations. 
\textbf{First}, our synthetic pre-trained models (4th and 5th row) already have strong generalization ability, which can be directly applied to realistic data without fine-tuning. Surprisingly, the model trained on RAD-S can even outperform the state-of-the-art B$^3$~\cite{Brick} by 0.8\% in mIoU. 
\textbf{Second}, following~\cite{LEGO,Brick}, we fine-tune our pre-trained model on each class of MNIST-C dataset, \ie, class-specific model (3rd row), in an unsupervised manner. Our method outperforms B$^3$~\cite{Brick} by 8.1\% in mIoU.
\textbf{Finally}, compared with reinforcement learning based GNN, our proposed tree-transformer is class-agnostic,
hence we can fine-tune our pre-trained model on all the classes jointly, \ie, class-agnostic models (6th and 7th row). The class-agnostic model can achieve a mIoU of 63.9\%, which is comparable with class-specific models. Our transfer learning based tree-transformer leads to a new state-of-the-art performance in both class-specific and class-agnostic modes.

\vspace{-1mm}
\paragraph{ModelNet3-C \& ModelNet40-C.} 
In Tab.~\ref{Tab:ModelNet} we compare our models under different settings with baselines on ModelNet3-C. We observe that: 
\textbf{First}, without any realistic data, our synthetic pre-trained model (5th row) surpasses state-of-the-art B$^3$~\cite{Brick} by a large margin (4.8\% in mIoU), which demonstrates the strong generalization ability of our tree-transformer.
\textbf{Second}, our model further yields an improvement of 7.6\% and 6.5\% in mIoU
with unsupervised class-specific and class-agnostic fine-tuning (4th and 6th row). Note that B$^3$~\cite{Brick} is a class-specific model that needs to be trained separately for each class, and our class-agnostic model exceeds it by 11.3\% in mIoU.
\textbf{Furthermore}, we report the results on ModelNet3-C and ModelNet40-C in Tab.~\ref{Tab:MoreModelNet}. Remarkably, our model fine-tuned on 3 classes can achieve a mIoU of 45.1\%, which is only 0.3\% lower than the model trained on the whole dataset.

\begin{table}[t]
\centering
\small
\setlength{\tabcolsep}{0.18cm}
\caption{\textcolor{black}{Comparison with the baselines on the RAD-1k dataset. We show the quantitative results including both IoU of assembled shape and accuracy (Acc) of predicted assembly actions. }}
\vspace{-3mm}
\label{Tab:RAD}
\begin{tabular}{l|c|ccc}
\toprule
Method & Supervision & IoU $\uparrow$ & Acc $\uparrow$ & Train Time (h) $\downarrow$ \\
\midrule
B$^3$~\cite{Brick} & Voxel & $42.0$ & ${32.0}$ & $27.0$ \\
BO~\cite{LEGO} & Voxel & $\underline{52.0}$  & $\underline{44.5}$ & $-$ \\
SL~\cite{LEGO_graph} & Action & $23.0$ & ${38.5}$ & $\underline{14.0}$ \\
Ours & Action & $\textbf{71.2}$ & $\textbf{63.2}$ & $\textbf{3.5}$ \\
\bottomrule
\end{tabular}
\vspace{-2mm}
\end{table}

\vspace{-1mm}
\paragraph{RAD.}
We further compare our models with BO~\cite{LEGO} and B$^3$~\cite{Brick} on the synthetic RAD dataset in Tab.~\ref{Tab:RAD}. 
Given the same supervision, our model outperforms SL~\cite{LEGO_graph} by 48.2\% in mIoU 
demonstrating the effectiveness of our tree-transformer and tree augmentation.
Our method also surpasses the SL by 24.7\% in terms of the accuracy of assembly action prediction, showing that our method works better on the task of assembly action prediction.
Moreover, our model is 8x faster than the voxel-based B$^3$~\cite{Brick} while still outperforming it by a large margin. 
Benefiting from parallel tree-transformer as well as the small search space modeled by \lego-Tree, our model achieves significant improvement in both performance and efficiency.

\begin{table}[t]
\centering{
\small
\setlength{\tabcolsep}{0.18cm}
\caption{\textcolor{black}{Ablation studies on ModelNet3-C. ``SP'' and ``AR'' denote synthetic pre-training and action reordering, respectively. $E_{geo}$ and $E_{str}$ are the global geometry embedding and tree structure embedding in the brick query.
}}
\vspace{-3mm}
\label{Tab:ablation}
\begin{tabular}{cccc|ccc|c}
\toprule
\multicolumn{4}{c|}{Method} & \multicolumn{3}{c|}{IoU} & \multirow{2}{*}{mIoU} \\
\cmidrule{1-7}
SP & $E_{str}$ & $E_{geo}$ & AR & \textit{airplane} & \textit{monitor} & \textit{table} & \\
\midrule
\textcolor{dark-red}{\xmark} & \cmark & \cmark & \textcolor{dark-red}{\xmark} & $32.1$ & $32.4$ & $15.6$ & $26.7$ \\
\cmark & \textcolor{dark-red}{\xmark}& \cmark & \cmark & $35.5$ & $55.7$ & $15.1$ & $35.4$ \\
\cmark & \cmark & \textcolor{dark-red}{\xmark}& \cmark & $29.6$ & $50.8$ & $16.0$ & $32.1$ \\
\cmark & \cmark & \cmark & \textcolor{dark-red}{\xmark} & $33.5$ & $54.7$ & $15.0$ & $34.3$ \\
\cmark & \cmark & \cmark & \cmark & $\textbf{42.2}$ & $\textbf{59.5}$ & $\textbf{23.4}$ & $\textbf{41.6}$ \\
\bottomrule
\end{tabular}}
\vspace{-3mm}
\end{table}

\subsection{Ablation Studies}
We first conduct ablation studies on the ModelNet3-C dataset in Tab.~\ref{Tab:ablation} to verify the effectiveness of our transfer learning strategy and components in tree-transformer queries. 
Then, we illustrate the choice of model hyper-parameter which shows consistency across three datasets.

\paragraph{Proposed Modules.} To investigate the effectiveness of our proposed modules, we conduct several ablation studies on the ModelNet3-C dataset. As shown in the 1st row in Tab.~\ref{Tab:ablation}, we train the tree-transformer directly on the realistic dataset without synthetic pre-training (SP). Action reordering is not adopted due to the lack of action labels. The results demonstrate that strong step-wise supervision in synthetic data can help to learn generalized knowledge for further fine-tuning.
The results in the 2nd and 3rd rows demonstrate the importance of both tree structure embedding $E_{str}$ and global geometry embedding $E_{geo}$. 
As the 4th row shown, action reordering (AR) plays an important role in enhancing the generalization ability by narrowing the domain between the ordered synthetic data and unordered realistic data by changing the root brick of the tree. 

\begin{table}[t]
\centering
\small
\caption{\textcolor{black}{Ablation study of different brick types and brick numbers on ModleNet3-C.}}
\vspace{-3mm}
\label{Tab:Brick_type}
\begin{tabular}{c|c|c|c|c}
\toprule
Primitives & \textit{airplane} & \textit{monitor} & \textit{table} & mIoU \\
\midrule
$\backsim60$ Brick(2,4) & ${44.6}$ & ${59.9}$ & ${23.5}$ & ${42.7}$ \\
$\backsim200$ Brick(2,4) & ${46.7}$ & ${60.2}$ & ${24.8}$ & ${43.9}$ \\
$\backsim200$ Brick(1,2) & ${57.0}$ & ${65.8}$ & ${31.2}$ & ${51.3}$ \\
$\backsim500$ Brick(1,2) & $\textbf{59.1}$ & $\textbf{70.6}$ & $\textbf{34.3}$ & $\textbf{54.7}$ \\
\bottomrule
\end{tabular}
\end{table}

\begin{figure*}[t]
\centering
\vspace{-1mm}
\includegraphics[width=.79\linewidth]{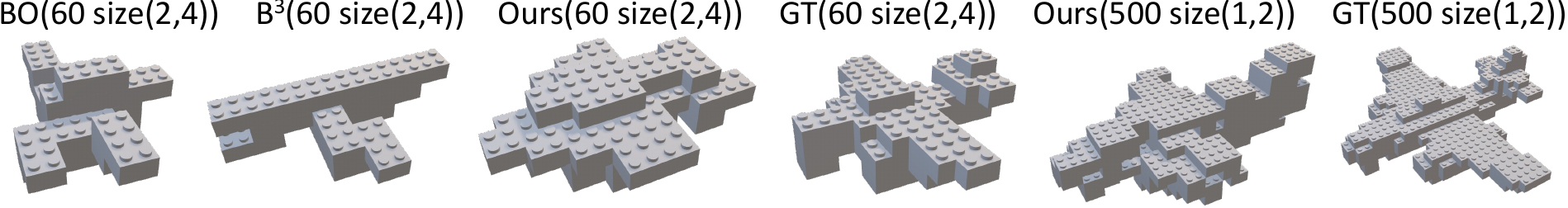}
\vspace{-4mm}
\caption{Visual comparison of baselines, brick number, brick size.}
\vspace{-2mm}
\label{fig:vis_comp}
\end{figure*}

\paragraph{Brick Type and Number.}
We further analyze the generalizability of the brick type and brick number of our model on the ModelNet3-C dataset. We conduct the experiments on four different basic primitive settings, using two types Brick(2, 4), Brick(1, 2), and three number settings $60$, $200$, $500$. As shown in Tab.~\ref{Tab:Brick_type} and Fig.~\ref{fig:vis_comp}, our method achieves comparable results across all kinds of settings, which proves it is easily expanded to more number of bricks and different types of bricks. Notably, our model improves the results with the more basic brick Brick(1, 2) or more bricks, which shows that our model has the potential to assemble objects with more details.
For a fair comparison, we use about $60$ Brick(2, 4) in the main experiment, which is the same setting used in baselines.

\begin{figure*}[t]
\centering
\includegraphics[width=.99\linewidth]{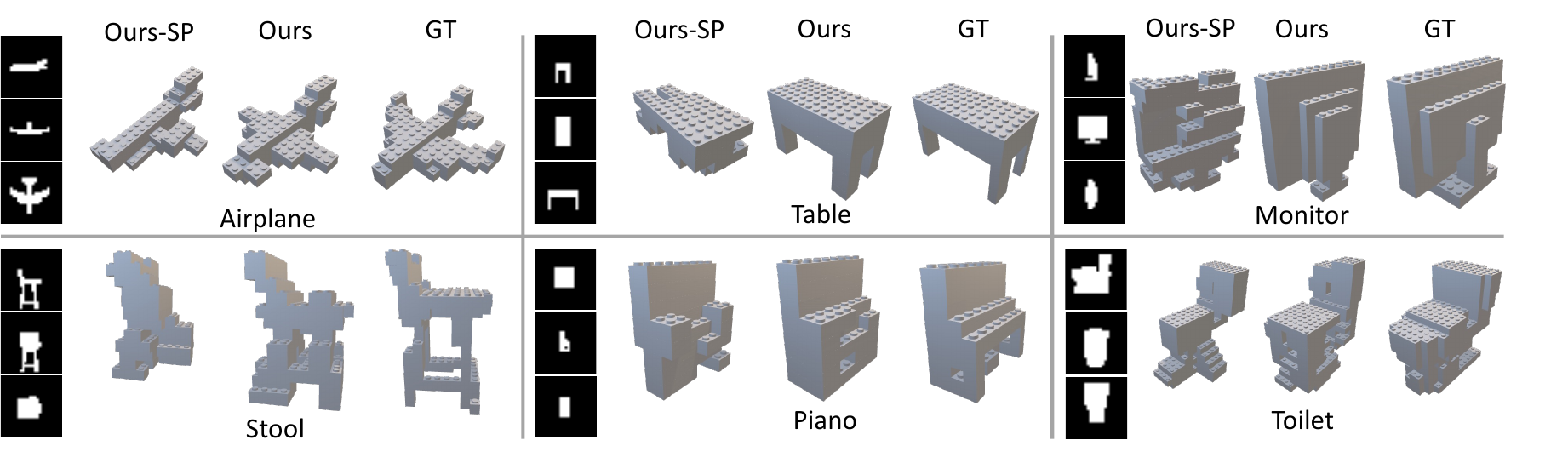}
\vspace{-5mm}
\caption{Visualization of the assembled objects in the ModelNet40-C.
`Ours-SP' is the result of the model only with pre-training.}
\vspace{-2mm}
\label{fig:vis_main}
\end{figure*}

\subsection{Visualization}
In Fig.~\ref{fig:vis_main}, we show the qualitative results of our synthetic pre-trained (Ours-SP) and action-to-silhouette fine-tuned (Ours) models on the ModelNet40-C datasets. 
\textbf{First}, our pre-trained model has a strong generalization ability that can assemble objects from unseen classes. \textbf{Second}, with silhouette self-training, our class-agnostic model adapts to real objects with different structures from the synthetic data. 
All the observations demonstrate the effectiveness of the synthetic-to-real transfer learning and the generalizability of the tree-transformer.

\begin{figure}[t]
\centering
\includegraphics[width=.55\linewidth]{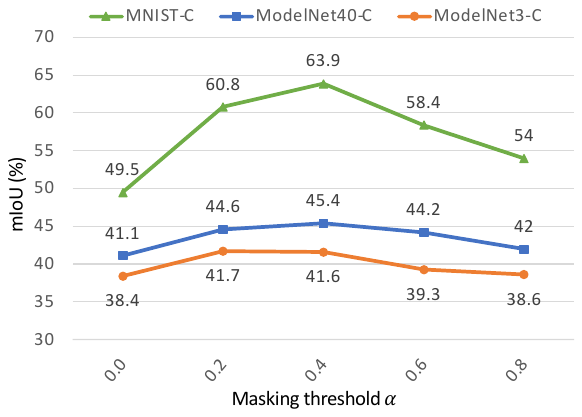}
\vspace{-3mm}
\caption{Parameter analysis of voxelization masking threshold $\alpha$ on different datasets.}
\vspace{-2mm}
\label{fig:param}
\end{figure}

\subsection{Parameter Analysis.}
\vspace{-1mm}
We further analyze the sensitivity of our model to the important hyper-parameter of voxelization masking threshold $\alpha$ (\cf Eqn.~\eqref{eq:map+sum+mask}) on MNIST-C, ModelNet3-C, and ModelNet40-C.  Fig.~\ref{fig:param} shows the performance first increases along with $\alpha$ and peaks when $\alpha=0.4$ since more voxels with low confidence and incorrect occupancy get discarded.
However, as $\alpha$ continues to increase, more voxels are filtered out thus leading to the lack of supervision for plenty of actions. As a result, the performance drops by a considerable margin of 9.9\%, 3.0\%, and 3.4\% in mIoU on MNIST-C, ModelNet3-C, and ModelNet40-C, respectively.

\section{Conclusion}
We have proposed a novel framework for the Sequential Brick Assembly (SBA) task. We design a breadth-first search (BFS) \lego-Tree to represent the sequential assembly actions and then build a class-agnostic Tree-Transformer for the task.
To mitigate the lack of ground truth action labels in real datasets, we leverage synthetic-to-real transfer learning. The model is pre-trained on synthetic data with full action supervision, and further fine-tuned on real data with self-supervision on image silhouettes obtained from our proposed action-to-silhouette projection. Experiments on three realistic benchmarks show that our framework achieves state-of-the-art performance on SBA even without any annotation in the target dataset.

\paragraph{Acknowledgement.} 
This research/project is supported by the National Research Foundation Singapore and DSO National Laboratories under the AI Singapore Programme (Award Number: AISG2-RP-2020-016), and the Tier 2 grant MOE-T2EP20120-0011 from the Singapore Ministry of Education.

%
%

\bibliographystyle{IEEEtran}
\bibliography{main}

\begin{thebibliography}{10}
\providecommand{\url}[1]{\texttt{#1}}
\providecommand{\urlprefix}{URL }
\providecommand{\doi}[1]{https://doi.org/#1}

\bibitem{arsalan2017synthesizing}
Arsalan~Soltani, A., Huang, H., Wu, J., Kulkarni, T.D., Tenenbaum, J.B.: Synthesizing 3d shapes via modeling multi-view depth maps and silhouettes with deep generative networks. In: CVPR (2017)

\bibitem{brochu2010tutorial}
Brochu, E., Cora, V.M., De~Freitas, N.: A tutorial on bayesian optimization of expensive cost functions, with application to active user modeling and hierarchical reinforcement learning. In: arXiv preprint arXiv:1012.2599 (2010)

\bibitem{chang2015shapenet}
Chang, A.X., Funkhouser, T., Guibas, L., Hanrahan, P., Huang, Q., Li, Z., Savarese, S., Savva, M., Song, S., Su, H., et~al.: Shapenet: An information-rich 3d model repository. In: arXiv preprint arXiv:1512.03012 (2015)

\bibitem{chen2022neural}
Chen, Y.C., Li, H., Turpin, D., Jacobson, A., Garg, A.: Neural shape mating: Self-supervised object assembly with adversarial shape priors. In: CVPR (2022)

\bibitem{choy20163d}
Choy, C.B., Xu, D., Gwak, J., Chen, K., Savarese, S.: 3d-r2n2: A unified approach for single and multi-view 3d object reconstruction. In: ECCV (2016)

\bibitem{Brick}
Chung, H., Kim, J., Knyazev, B., Lee, J., Taylor, G.W., Park, J., Cho, M.: Brick-by-brick: Combinatorial construction with deep reinforcement learning. In: NeurIPS (2021)

\bibitem{curless1996volumetric}
Curless, B., Levoy, M.: A volumetric method for building complex models from range images. In: SIGGRAPH (1996)

\bibitem{debevec1996modeling}
Debevec, P.E., Taylor, C.J., Malik, J.: Modeling and rendering architecture from photographs: A hybrid geometry-and image-based approach. In: SIGGRAPH (1996)

\bibitem{dosovitskiy2020vit}
Dosovitskiy, A., Beyer, L., Kolesnikov, A., Weissenborn, D., Zhai, X., Unterthiner, T., Dehghani, M., Minderer, M., Heigold, G., Gelly, S., Uszkoreit, J., Houlsby, N.: An image is worth 16x16 words: Transformers for image recognition at scale. ICLR  (2021)

\bibitem{eilers2016lego}
Eilers, S.: The lego counting problem. The American Mathematical Monthly  (2016)

\bibitem{fan2017point}
Fan, H., Su, H., Guibas, L.J.: A point set generation network for 3d object reconstruction from a single image. In: CVPR (2017)

\bibitem{gadelha2020learning}
Gadelha, M., Gori, G., Ceylan, D., Mech, R., Carr, N., Boubekeur, T., Wang, R., Maji, S.: Learning generative models of shape handles. In: CVPR (2020)

\bibitem{ghosh2021generalization}
Ghosh, D., Rahme, J., Kumar, A., Zhang, A., Adams, R.P., Levine, S.: Why generalization in rl is difficult: Epistemic pomdps and implicit partial observability. NeurIPS  (2021)

\bibitem{gower1998lego}
Gower, R., Heydtmann, A., Petersen, H.: Lego: Automated model construction  (1998)

\bibitem{LiJiang2018GALGA}
Jiang, L., Shi, S., Qi, X., Jia, J.: Gal: Geometric adversarial loss for single-view 3d-object reconstruction. In: ECCV (2018)

\bibitem{kar2017learning}
Kar, A., H{\"a}ne, C., Malik, J.: Learning a multi-view stereo machine. In: NeurIPS (2017)

\bibitem{LEGO}
Kim, J., Chung, H., Lee, J., Cho, M., Park, J.: Combinatorial 3d shape generation via sequential assembly. In: NeurIPS Workshop (2020)

\bibitem{lee2015finding}
Lee, S., Kim, J., Kim, J.W., Moon, B.R.: Finding an optimal lego{\textregistered} brick layout of voxelized 3d object using a genetic algorithm. In: The Genetic and Evolutionary Computation Conference (GECCO) (2015)

\bibitem{lennon2021image2lego}
Lennon, K., Fransen, K., O'Brien, A., Cao, Y., Beveridge, M., Arefeen, Y., Singh, N., Drori, I.: Image2lego: Customized lego set generation from images. arXiv preprint arXiv:2108.08477  (2021)

\bibitem{li2020learning}
Li, Y., Mo, K., Shao, L., Sung, M., Guibas, L.: Learning 3d part assembly from a single image. In: ECCV (2020)

\bibitem{li2018learning}
Li, Y., Vinyals, O., Dyer, C., Pascanu, R., Battaglia, P.: Learning deep generative models of graphs. arXiv preprint arXiv:1803.03324  (2018)

\bibitem{mandikal20183d}
Mandikal, P., Navaneet, K., Agarwal, M., Babu, R.V.: 3d-lmnet: Latent embedding matching for accurate and diverse 3d point cloud reconstruction from a single image. In: BMVC (2018)

\bibitem{mildenhall2020nerf}
Mildenhall, B., Srinivasan, P.P., Tancik, M., Barron, J.T., Ramamoorthi, R., Ng, R.: Nerf: Representing scenes as neural radiance fields for view synthesis. In: ECCV (2020)

\bibitem{park2019deepsdf}
Park, J.J., Florence, P., Straub, J., Newcombe, R., Lovegrove, S.: Deepsdf: Learning continuous signed distance functions for shape representation. In: CVPR (2019)

\bibitem{paschalidou2019superquadrics}
Paschalidou, D., Ulusoy, A.O., Geiger, A.: Superquadrics revisited: Learning 3d shape parsing beyond cuboids. In: CVPR (2019)

\bibitem{GernotRiegler2016OctNetLD}
Riegler, G., Osman~Ulusoy, A., Geiger, A.: Octnet: Learning deep 3d representations at high resolutions. In: CVPR (2017)

\bibitem{schulman2017proximal}
Schulman, J., Wolski, F., Dhariwal, P., Radford, A., Klimov, O.: Proximal policy optimization algorithms. arXiv preprint arXiv:1707.06347  (2017)

\bibitem{ZaiShi20213DRETRES}
Shi, Z., Meng, Z., Xing, Y., Ma, Y., Wattenhofer, R.: 3d-retr: End-to-end single and multi-view 3d reconstruction with transformers. In: BMVC (2021)

\bibitem{snavely2006photo}
Snavely, N., Seitz, S.M., Szeliski, R.: Photo tourism: exploring photo collections in 3d. In: SIGGRAPH (2006)

\bibitem{XingyuanSun2018Pix3DDA}
Sun, X., Wu, J., Zhang, X., Zhang, Z., Zhang, C., Xue, T., Tenenbaum, J.B., Freeman, W.T.: Pix3d: Dataset and methods for single-image 3d shape modeling. In: CVPR (2018)

\bibitem{tatarchenko2017octree}
Tatarchenko, M., Dosovitskiy, A., Brox, T.: Octree generating networks: Efficient convolutional architectures for high-resolution 3d outputs. In: ICCV (2017)

\bibitem{LEGO_graph}
Thompson, R., Ghalebi, E., DeVries, T., Taylor, G.W.: Building lego using deep generative models of graphs. In: NeurIPS Workshop (2020)

\bibitem{tulsiani2018multi}
Tulsiani, S., Efros, A.A., Malik, J.: Multi-view consistency as supervisory signal for learning shape and pose prediction. In: CVPR (2018)

\bibitem{tulsiani2018factoring}
Tulsiani, S., Gupta, S., Fouhey, D.F., Efros, A.A., Malik, J.: Factoring shape, pose, and layout from the 2d image of a 3d scene. In: CVPR (2018)

\bibitem{tulsiani2017learning}
Tulsiani, S., Su, H., Guibas, L.J., Efros, A.A., Malik, J.: Learning shape abstractions by assembling volumetric primitives. In: CVPR (2017)

\bibitem{vaswani2017attention}
Vaswani, A., Shazeer, N., Parmar, N., Uszkoreit, J., Jones, L., Gomez, A.N., Kaiser, {\L}., Polosukhin, I.: Attention is all you need. In: NeurIPS (2017)

\bibitem{walsman2022break}
Walsman, A., Zhang, M., Kotar, K., Desingh, K., Farhadi, A., Fox, D.: Break and make: Interactive structural understanding using lego bricks. ECCV  (2022)

\bibitem{wang2022translating}
Wang, R., Zhang, Y., Mao, J., Cheng, C.Y., Wu, J.: Translating a visual lego manual to a machine-executable plan. ECCV  (2022)

\bibitem{wu2017marrnet}
Wu, J., Wang, Y., Xue, T., Sun, X., Freeman, B., Tenenbaum, J.: Marrnet: 3d shape reconstruction via 2.5 d sketches. In: NeurIPS (2017)

\bibitem{wu2016learning}
Wu, J., Zhang, C., Xue, T., Freeman, B., Tenenbaum, J.: Learning a probabilistic latent space of object shapes via 3d generative-adversarial modeling. In: NeurIPS (2016)

\bibitem{wu2023leveraging}
Wu, R., Tie, C., Du, Y., Zhao, Y., Dong, H.: Leveraging se (3) equivariance for learning 3d geometric shape assembly. In: ICCV (2023)

\bibitem{wu2020pq}
Wu, R., Zhuang, Y., Xu, K., Zhang, H., Chen, B.: Pq-net: A generative part seq2seq network for 3d shapes. In: CVPR (2020)

\bibitem{wu20153d}
Wu, Z., Song, S., Khosla, A., Yu, F., Zhang, L., Tang, X., Xiao, J.: 3d shapenets: A deep representation for volumetric shapes. In: CVPR (2015)

\bibitem{xie2019pix2vox}
Xie, H., Yao, H., Sun, X., Zhou, S., Zhang, S.: Pix2vox: Context-aware 3d reconstruction from single and multi-view images. In: ICCV (2019)

\bibitem{xu2023unsupervised}
Xu, X., Guerrero, P., Fisher, M., Chaudhuri, S., Ritchie, D.: Unsupervised 3d shape reconstruction by part retrieval and assembly. In: CVPR (2023)

\bibitem{zakka2020form2fit}
Zakka, K., Zeng, A., Lee, J., Song, S.: Form2fit: Learning shape priors for generalizable assembly from disassembly. In: ICRA (2020)

\bibitem{zhan2020generative}
Zhan, G., Fan, Q., Mo, K., Shao, L., Chen, B., Guibas, L.J., Dong, H., et~al.: Generative 3d part assembly via dynamic graph learning. NeurIPS  (2020)

\bibitem{zou20173d}
Zou, C., Yumer, E., Yang, J., Ceylan, D., Hoiem, D.: 3d-prnn: Generating shape primitives with recurrent neural networks. In: ICCV (2017)

\end{thebibliography}


\clearpage






\appendix
\renewcommand\thefigure{S\arabic{figure}}
\renewcommand\thetable{S\arabic{table}}

\section{Our Method}
\subsection{Definition Details}
The connection type between two \lego bricks can be divided into four categories including x-axis offset, y-axis offset, z-axis offset, and relative rotation. We define the z-axis offset separately as direction type $dir_n$ and the remaining as tree connection type $t_n$. Exiting works~\cite{LEGO, Brick} represent the connection type between any two \lego bricks as an action. However, this leads to an excessively large search space when constructing a 3D object with tens of bricks. To reduce the search space, we redefine an action as the connection between the current brick with all its neighbors instead of any two bricks. We denote the action type as $a_n$.

\begin{figure}
    \centering
    \includegraphics[width=.95\linewidth]{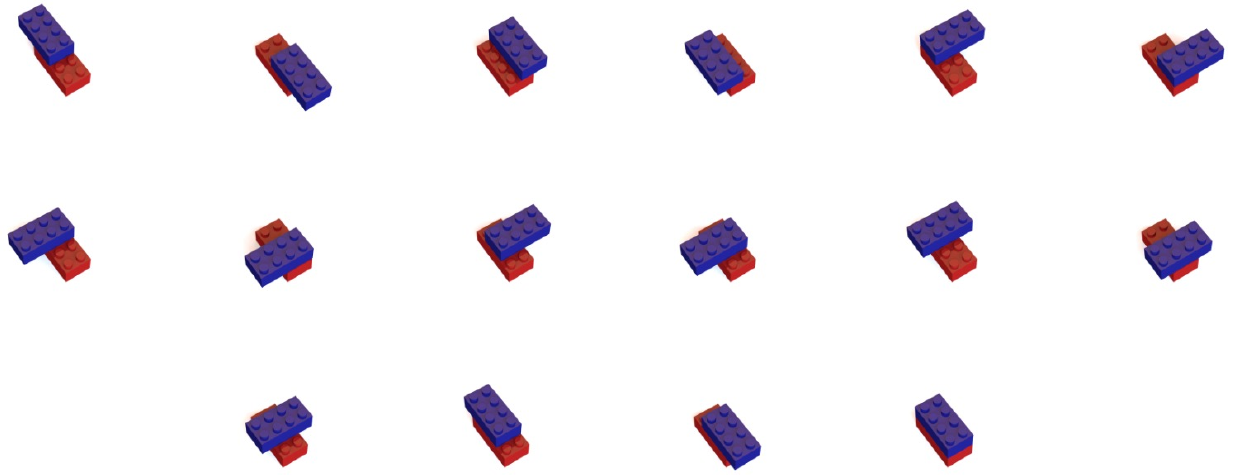}
    \caption{Illustration of tree connection type. There are 16 different types between two 2$\times$4 \lego bricks. Figure from~\cite{LEGO}.}
\label{fig:connection_type}
\end{figure}

\paragraph{Direction Type.}
The direction type is denoted by $dir_n \in \{0,1\}$, where $0$ represents that the brick $b_n$ is on top of its parent brick, and $1$ indicates the opposite direction, \ie at the bottom of its parent brick. For example, the direction type of the blue brick in Fig.~\ref{fig:connection_type} is $dir_{blue}=0$ when taking the red brick as the parent brick.

\paragraph{Tree Connection Type.}
The tree connection type $t_n\in \{0,1,...,T\}$ encodes the information on the x-axis offset, y-axis offset, and relative rotation. We follow the definition in ~\cite{LEGO, Brick} and show all possible 16 tree connection types between two $2\times 4$ 
\lego bricks in Fig.~\ref{fig:connection_type}.

\paragraph{Action Type.}
Each action can be divided into two sub-actions based on the direction type $a_n=[a_n^{up}, a_n^{down}]$, where $a_n^{up}$ ($a_n^{down}$) represents the top (bottom) of the current brick. Each sub-action represents the connection type between the current brick to all its neighboring bricks. Take the $2\times 4$ \lego brick for example, given the constraint that at least 4 of all 8 studs need to overlap with each other, there are 17 types of connection between any two bricks when the direction type is fixed, including 16 types of single-neighbor shown in Fig.~\ref{fig:connection_type} and one type of no neighbor. In total, there are 34 types of sub-actions, \ie, $a_n^{up}, a_n^{down} \in \{0,1,...,33\}$.

\subsection{Connection Order}
Our model predicts connections of the brick $b_n$ and all its neighbors in one shot (noted as $a_n$) to reduce the computational costs. These neighbors are added to the \lego-Tree in order following two rules: 1) The upper brick first($a_n^{up}$ first, then $a_n^{down}$). 2) The smaller tree connection type first. (Neighbor with $t_n'=3$ is before Neighbor with $t_n''=4$.)
Taking Fig.~3(a) as an example, the model first predicts that $b_0$ has two neighbors and they are both on the upper side $a_0^{up}$. It chooses the neighbor with a smaller tree connection type as $b_1$ and the other as $b_2$. 
Then, the model predicts neighbors $b_3$ (in $a_1^{up}$), $b_4$ (in $a_1^{down}$) for $b_1$ according to the breadth-first scheme. Notably, the predicted $b_1$-$b_0$ connection is discarded due to illegal connections colliding with existing bricks $b_0$.
Finally, the model predicts neighbor $b_5$ for $b_2$ and discards illegal connections $b_2$-$b_0$ and $b_2$-$b_1$. The rest of the predicted connections are illegal and dropped.

\subsection{Action Reordering}
As explained in Sec.~3.3 of the main paper, we do not assume that the root brick is located at the bottom of an object for realistic data. There are three reasons: \textbf{First}, the bottom center of a real 3D object may not fit a \lego brick, or connects with other parts with very thin structure, \eg the base of a table. In this case, the model tends to predict the stop action for the first few steps, which results in inaccurate assembly. \textbf{Second}, the depth of the \lego-Tree gets larger when the root brick is at the bottom. This increases the complexity of our tree-based representation, which leads to unstable predictions. \textbf{Third}, the object only appears in the upper half of the image when the root brick is at the center of an image, and therefore cannot make full use of the image encoder. In view of this, we utilize the action reordering strategy to bridge the gap between the synthetic and realistic data.

\begin{table}[t]
\centering
\small
\caption{\textcolor{black}{Implementation Details of our ViT~\cite{dosovitskiy2020vit}-based image encoder in tree-transformer.}}
\setlength{\tabcolsep}{0.5cm}
\begin{tabular}{cc|cc}
\toprule
Parameters & Value & Parameters & Value \\
\midrule
Layer/Depth & $4$ & Channel & $3$ \\
MLP Dim & $512$ & Head & $8$ \\
Feature Dim & $64$ & CLS Token & No \\
Image Size & $64$ & Patch Size & $16$ \\
\bottomrule
\end{tabular}
\label{Tab:ViT_param}
\end{table}

\section{Experiments}

\subsection{Implementation Details.}
We use the ViT~\cite{dosovitskiy2020vit}-based image encoder with some modifications as shown in Tab.~\ref{Tab:ViT_param}. The input multi-view silhouettes are concatenated in the channel dimension to obtain $I \in R^{C\times H\times W}$, where $C=3, H=W=64$. The input image is divided into 16 patches, where each patch has a shape of $R^{3\times 16 \times 16}$. The dimension of the output image features is $R^{16\times 64}$.
The brick decoder consists of a four-layer transformer decoder with a feature dimension of 64 and a brick embedding module with an embedding dimension of 64.
The action head consists of two 64$\times$34 linear layers, which predicts two tree connection types $[a_n^{up},a_n^{down}]$.

\begin{table}[t]
\centering
\small
\caption{\textcolor{black}{Ablation study of different backbone for encoder and decoder on RAD-1k dataset.}}
\vspace{-3mm}
\label{Tab:Transformer}
\begin{tabular}{c|c|c|c}
\toprule
Enc.+Dec. & TF+RNN & ResNet+TF & TF+TF \\
\midrule
IoU & ${56.3}$ & ${67.7}$ & $\textbf{71.2}$ \\
\bottomrule
\end{tabular}
\vspace{-2mm}
\end{table}

\subsection{Ablation Studies}
\paragraph{Backbone.}
We conduct the experiment on the RAD-1k dataset to verify the effectiveness of the transformer-based architecture, where we replace the transformer encoder with ResNet (ResNet+TF) or the decoder with RNN (TF+RNN). The results in Tab.~\ref{Tab:Transformer} show the performance drops by 3.5\% when using ResNet as the encoder and 14.9\% when using RNN as the decoder.
The advantages of applying the tree-transformer to the SBA task are two-fold: 1) The multi-head attention mechanism inherently learns the relationship between bricks, and this aggregation of context information is important for predicting the action for each brick.
2) We modify the original position embedding to include both tree structure embedding and global geometry embedding, which encodes both local and global information of each brick. 
In contrast, no sequential information is perceived in CNN, and the strong correlations between sequentially distant bricks are missing in RNN.

\subsection{Comparison on Computational Complexity}

\begin{table}[t]
\centering
\small
\vspace{-4mm}
\caption{{\textcolor{black}{Comparison of inference time, model size, primitive number. BO does not have a `model' or any learnable parameters. }}}
\setlength{\tabcolsep}{0.7mm}
\begin{tabular}{c|ccccc}
\hline
Method (Num) & SQ(20) & $B^3$(60) & BO(60) & SL(60) & Ours(60-500) \\
\hline
Inf. Time (s) $\downarrow$ & $\textbf{0.02}$ & $0.28$ & $1.00$ & $\underline{0.15}$ & $\textbf{0.02}$ \\
Mem (MB) $\downarrow$ & $50$ & $31$ & $-$ & $\underline{10}$ & $\textbf{8.4}$ \\
\hline
\end{tabular}
\label{Tab:Time}
\end{table}
The computational complexity comparison is shown in Tab.~\ref{Tab:Time}. 
Benefiting from the parallel tree-transformer, our model achieves significant efficiency improvements and maintains the same inference time even as the brick number increases.
Moreover, our framework achieves notably better results with fewer parameters.

\subsection{Comparison on Pre-training}

\begin{table}[t]
\centering
\small
\caption{{\textcolor{black}{Comparison of pre-train (PT) strategy. SQ and BO are per-object optimization methods and are not suitable for PT. }}}
\setlength{\tabcolsep}{0.6mm}
\begin{tabular}{c|cccc}
\hline
Method(Supervision) & SQ (3D) & BO (3D) & $B^3$ (3D) & Ours (2D) \\
\hline
w/o pre-train & $29.2$ & $29.7$ & $\textbf{30.3}$ & $\underline{29.8}$ \\
w/ pre-train & $-$ & $-$ & $\underline{32.8}$ & $\textbf{42.7}$ \\
\hline
\end{tabular}
\label{Tab:pretrain}
\end{table}
We present the results with and without pre-training (PT) in Tab.~\ref{Tab:pretrain}. Our model, self-supervised by 2D images, achieves comparable performance with baselines supervised by 3D labels. With general actions learned from the PT strategy, our model significantly surpasses the baselines.

\subsection{Visualization}
We further show more qualitative results of our approach on the ModelNet3-C~\cite{wu20153d}, ModelNet40-C~\cite{wu20153d}, and MNIST-C~\cite{Brick} datasets in Fig.~\ref{fig:vis_ModelNet} and Fig.~\ref{fig:vis_MNIST}. We can see that our class-agnostic model is able to generate realistic objects that are close to the ground truth for different categories. Moreover, our model consistently performs well for the sequential assembly task across multiple datasets, demonstrating its generalization ability.

\subsection{Discussion on Category-agnostic Property}
The poor cross-category generalizability of existing approaches is a significant factor. Baseline methods are prone to overfitting because they rely on \textit{class-specific} information, such as object structures or step-wise actions. Additionally, methods using reinforcement learning require complex optimization for each category, which limits their generalization ability~\cite{ghosh2021generalization}.
In contrast, our framework addresses these issues by first learning general actions from synthetic action labels and then fine-tuning through image self-supervision. Our \textit{category-agnostic} feedforward neural framework - comprising an image encoder, brick queries, and a brick decoder - further enhances generalization across categories.

\section{Assembly Demo}
We have included a sequential brick assembly video `\textbf{Assembly\_Airplane. mp4}' for a better illustration of our SBA task and the power of our proposed method. In this video, we assemble an airplane object from the ModelNet40-C dataset with about 200 $2\times 4$ \lego bricks. The red brick represents the latest assembled brick while the green bricks are the previously assembled bricks. We can see that the bricks are sequentially assembled one by one and finally we get a well-assembled airplane object. The video proves our model can sequentially assemble complex 3d objects with plenty of simple \lego bricks.

\section{Limitation \& Failure Cases}
\begin{figure}
\centering
\includegraphics[width=.99\linewidth]{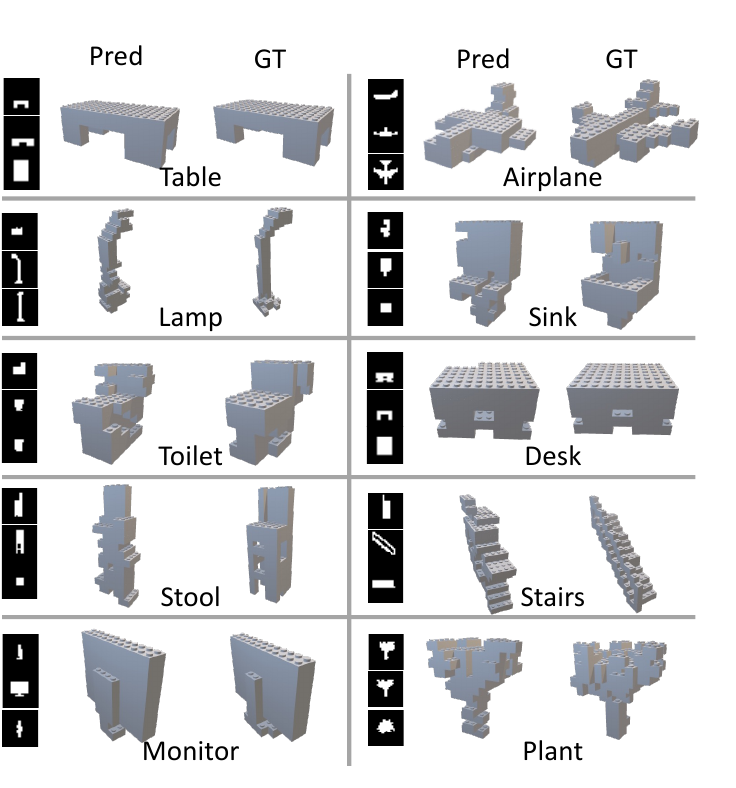}
\caption{Visualization of the assembled realistic objects from ten different categories. All the results use our class-agnostic model to generate the sequential assembly actions. }
\label{fig:vis_ModelNet}
\end{figure}

\begin{figure}
\centering
\includegraphics[width=.8\linewidth]{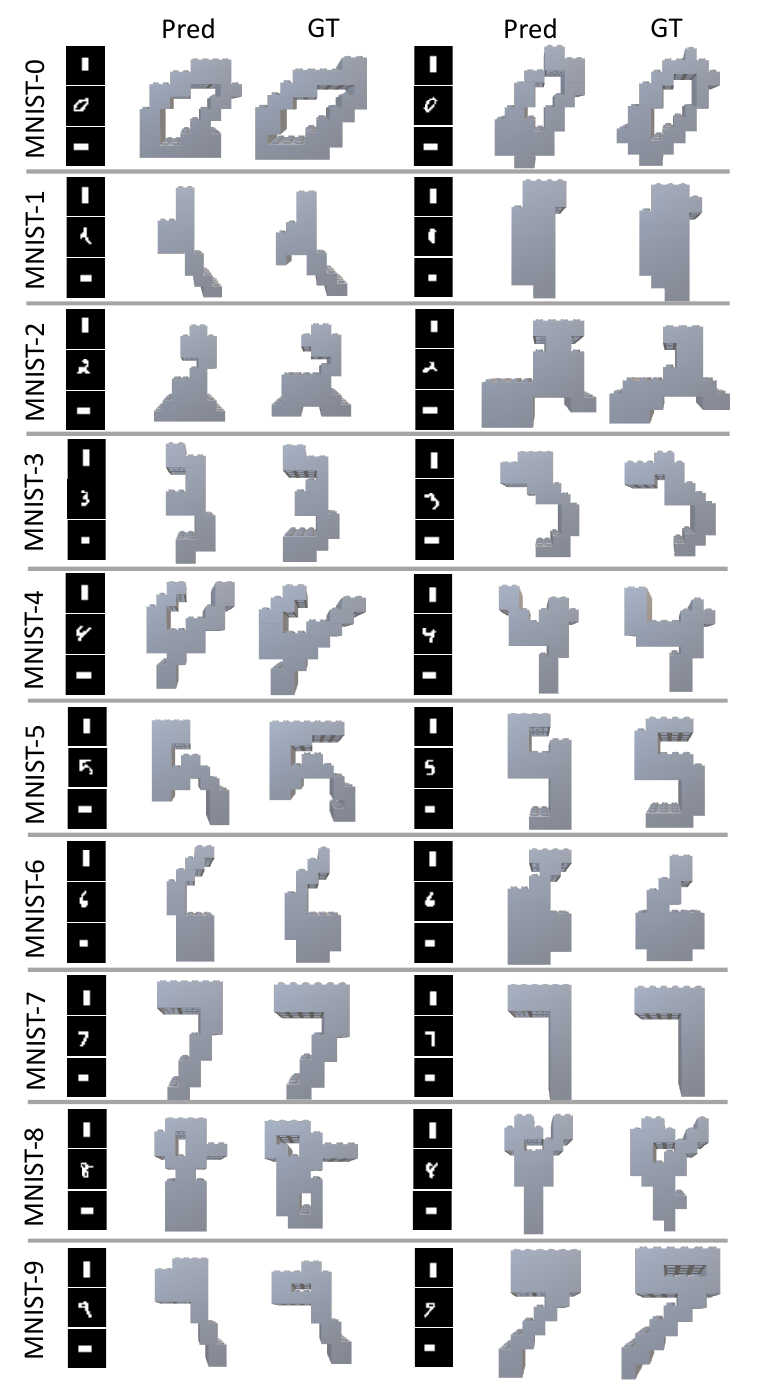}
\caption{Visualization of the assembled objects from MNIST-C. These objects belong to the categories from $\textit{MNIST-0}$ to $\textit{MNIST-9}$ (top to bottom), and two examples are shown for each category. All the results use our class-agnostic model to generate the sequential assembly actions. 
}
\label{fig:vis_MNIST}
\end{figure}

Although our method achieves significant improvement on both simple synthetic and complex realistic objects, we share the same limitation with previous works, \ie, unable to assemble some specific fine details of objects. As shown in the example \textit{Airplane} of the 1st row in Fig.~\ref{fig:vis_ModelNet}, our model can not accurately assemble the small tip of the airplane wing. Moreover, our method assembles the approximate outline of self-occlusion objects due to the limited observations, \ie, three view silhouettes, as the example \textit{Plant} of the last row in Fig.~\ref{fig:vis_ModelNet} shown.

\end{document}